\documentclass{bmvc2k}

\usepackage[linesnumbered,algoruled,boxed,lined]{algorithm2e}

\makeatletter 
\g@addto@macro{\@algocf@init}{\SetKwInOut{Parameter}{Parameters}} 
\makeatother

\usepackage{array}
\newcolumntype{C}[1]{>{\centering\arraybackslash}p{#1}}

\title{Case-Based Histopathological \\ Malignancy Diagnosis \\ using Convolutional Neural Networks}

\addauthor{Qicheng Lao}{qi_lao@encs.concordia.ca}{1}
\addauthor{Thomas Fevens}{fevens@encs.concordia.ca}{1}

\addinstitution{
	Department of Computer Science and Software Engineering,\\
	Concordia University,\\
	Montr\'{e}al, Qu\'{e}bec, Canada
}

\runninghead{Lao, Fevens}{Case-Based Histopathological Malignancy Diagnosis}

\def\etal{\emph{et al}\bmvaOneDot}

\begin{document}

\maketitle

\begin{abstract}
In practice, histopathological diagnosis of tumor malignancy often requires a human expert to scan through histopathological images at multiple magnification levels, after which a final diagnosis can be accurately determined. However, previous research on such classification tasks using convolutional neural networks primarily determine a diagnosis for a single magnification level. In this paper, we propose a case-based approach using deep residual neural networks for histopathological malignancy diagnosis, where a case is defined as a sequence of images from the patient at all available levels of magnification. Effectively, through mimicking what a human expert would actually do, our approach makes a diagnosis decision based on features learned in combination at multiple magnification levels. Our results show that the case-based approach achieves better performance than the state-of-the-art methods when evaluated on BreaKHis, a histopathological image dataset for breast tumors.
\end{abstract}

\section{Introduction}
\label{sec:intro}
Histopathology is regarded as the gold standard method for cancer diagnosis, including almost all types of cancers, such as breast, lung, colon and prostate cancer \cite{rubin2008rubin, gurcan2009histopathological, he2012histology}. Suspicious tissues are biopsied and the biopsy undergoes fixation, sectioning, and finally mounting on a slide. The biopsy section then is subjected to haematoxylin and eosin (H\&E) staining which is a routinely used staining procedure that enhances tissue structure and cell morphology. A pathologist would then thoroughly examine the H\&E stained slides under a microscope at multiple magnification levels, searching for morphological signatures indicating the onset or progression of cancerous tissues whose presence determines whether the tumor should be diagnosed as benign or malignant. The whole process, however, can be very time-consuming, since it is often required that the pathologist switch between magnification levels and jump among different image locations \cite{roa2010experimental}. In addition, the diagnosis from a pathologist can sometimes be subjective and heavily dependent on the experience of the pathologist \cite{he2012histology}.

In order to address the above problems, computer aided diagnosis (CAD) systems have been proposed to facilitate cancer diagnosis, not only to reduce labor work for the pathologist, but also to improve objectivity and consistency. Despite the work that has been done in the last few decades \cite{demir2005automated, gurcan2009histopathological, he2012histology, veta2014breast}, tumor malignancy classification remains still a challenge for most automatic cancer diagnosis applications due to the tremendous complexity of histopathological images, which can be due to various reasons including the staining variations in specimen treatment process \cite{mccann2015automated} and the diversity of tissue characteristics in different cancers. Therefore, a robust and reliable CAD system for cancer diagnosis has to be designed to capture all discriminative features in histopathological images effectively. However, as has been pointed out by many researchers \cite{demir2005automated, gurcan2009histopathological, he2012histology, spanhol2016breast}, when using traditional classification approaches, the feature engineering step can be very difficult that requires a fair amount of expert domain knowledge.

Recently, a wide variety of new deep learning technologies \cite{lecun2015deep, krizhevsky2012imagenet}, such as the convolutional neural network (CNN), first developed by LeCun \etal \cite{lecun1989backpropagation}, have achieved great success on various computer vision and pattern recognition tasks. Indeed, CNN has become the state-of-the-art method for image based classification problems, consistently outperforming traditional machine learning methods. More importantly, CNN can automatically extract discriminative features from images by itself. As a result, no hand-crafted feature engineering step is required anymore, which saves considerable efforts in most applications including histopathological image classification.

\section{Previous work}
Due to its superior performance compared to traditional machine learning methods, CNN has been widely applied to histopathological cancer diagnosis problems. Cire{\c{s}}an \etal \cite{cirecsan2013mitosis} use CNN to detect mitosis in breast cancer histological images and won the ICPR 2012 mitosis detection competition. Sirinukunwattana \etal \cite{sirinukunwattana2016locality} propose a spatially constrained CNN for nucleus detection and then a Neighboring Ensemble Predictor (NEP) coupled with CNN for nucleus classification in colon caner histological images, and achieve the highest average F1 score for this problem compared to other methods. Although both of the above two papers are not directly working on tumor malignancy classification, their results could undoubtedly benefit cancer diagnosis, since both mitosis and nuclear characteristics are important indicators for cancerous tissue detection. Direct work on malignancy classification have also been published. For example, Cruz-Roa \etal \cite{cruz2014automatic} show that a CNN classifier achieves a balanced accuracy of 84.23\% for the detection of invasive ductal carcinoma, where the best performance of methods using handcrafted features and classifiers is 78.74\%. Similarly, Litjens \etal \cite{litjens2016deep} also demonstrate that CNN improves the efficacy of prostate cancer diagnosis. 

We note that the previous work mentioned above on histopathological image classification using convolutional neural networks are done on whole slide images (WSI), and the patches used for training are extracted from the original images at a certain fixed magnification level. However, an experienced pathologist would not choose to determine a diagnosis decision based on a single magnification level. In practice, it is often required that the pathologists evaluate the histopathological slides at multiple magnification levels \cite{roa2010experimental, romo2014discriminant}, as different magnifications give different features. For instance, lower magnification gives global texture information and tissue structure while higher magnification resolve more on cellular morphology and sub-cellular details \cite{gurcan2009histopathological}. Sometimes it is difficult to determine a diagnosis merely based on a single magnification level. Only by integrating all the features at multiple magnification levels, a confident diagnosis can be determined.

Recently, an image dataset BreaKHis is released \cite{spanhol2016dataset}, which provides histopathological images of breast tumor at multiple magnification levels (40$\times$, 100$\times$, 200$\times$ and 400$\times$). Both traditional methods using handcrafted features \cite{spanhol2016dataset} and CNN method \cite{spanhol2016breast} have been applied on this dataset for malignancy classification, and it has been shown that by combining different CNNs using fusion rules, the CNN performance has an improvement of 6\% in classification accuracy, compared to traditional methods. However, one disadvantage of this paper \cite{spanhol2016breast}, is that four CNN classifiers have to be trained, with one classifier specialized for each of the four magnifications. Seeking to find a better solution to this problem, Bayramoglu \etal \cite{bayramogludeep} propose a magnification independent approach with both single-task (malignancy) and multi-task (malignancy and magnification) classification, where they ignore magnification information of the image and train a unique CNN classifier for all magnifications. Although the performance is slightly impaired, it indeed improves the efficiency. Nevertheless, when evaluated on the testing sets, both of the previous work \cite{spanhol2016breast, bayramogludeep} on BreaKHis dataset using CNN fail to determine a diagnosis for a patient based on features from multiple magnification levels at the same time. Instead, they give separate classification accuracy for each individual magnification, independent to other available magnifications.

\section{Histopathological case-based classification}
\label{sec:approach}
\begin{figure}[ht]
	\centering
	\includegraphics[scale=0.32]{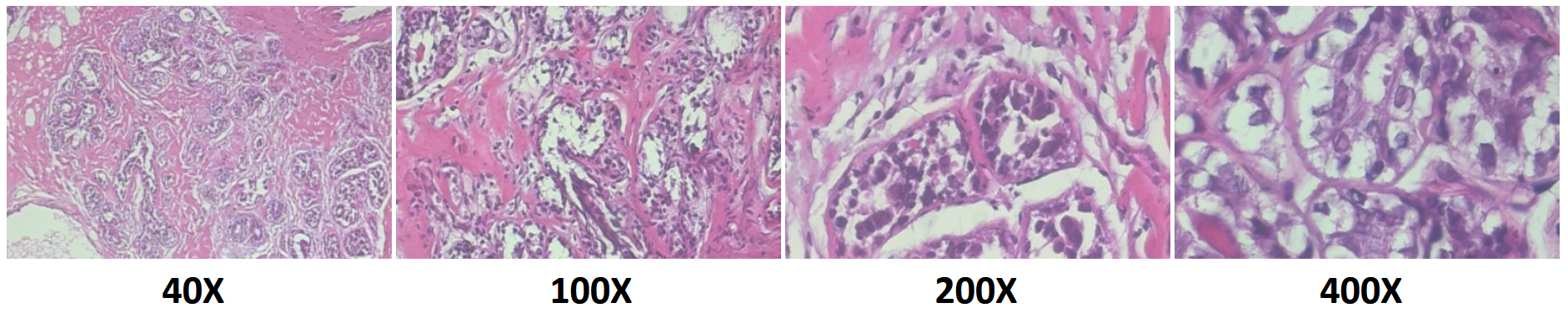}
	\caption{ A typical histopathological case of breast tumor with different magnifications. }
	\label{fig:showcaseimage}
\end{figure}

In order to build a more reasonable and reliable computer aided diagnosis system, we propose a case-based approach for histopathological malignancy classification, where a case is defined as a sequence of images including one or more images from each of all the available levels of magnification for a certain dataset. For example, for the BreaKHis dataset, a typical case could consist of one or more images at each of the following magnifications in order: 40$\times$, 100$\times$, 200$\times$ and 400$\times$ (Figure \ref{fig:showcaseimage}). A trained classifier should be able to learn all the features from different magnification images, and give a unique and more accurate result based on all information given (e.g. tissue structure at lower magnification, cell phenotype at higher magnification), equivalent to how an histopathological expert would choose to perform analysis at multiple magnification levels.

In this section, we first present our algorithm that constructs a case-based image set from any given histopathological image dataset with multiple magnifications and malignancies (Section \ref{sec:algorithm}). We then introduce a CNN model to classify our histopathological cases (Section \ref{sec:classifier}). Finally, we describe the three performance metrics that will be used for the evaluation of histopathological case-based classification (Section \ref{sec:metrics}).

\subsection{Case-based image set initialization}
\label{sec:algorithm}
Histopathological image datasets are often given as images in multiple separated magnifications, but not as cases. Therefore, the first step is to build an appropriate number of histopathological cases based on the given dataset. To limit the size of the input set, the cases will include exactly one image from each magnification level. Algorithm \ref{algo:casebuild} describes the initialization of a case-based image set from the original dataset with multiple magnifications and malignancies. Put simply, for each case build, the algorithm randomly chooses one image from each subset of images that belong to different magnifications, with the restriction that all images in the same case must have the same malignancy label, which will also be the final class label for the resulting case. For simplicity, we illustrate in Algorithm \ref{algo:casebuild}, assuming that two types of malignancy (benign and malignant) and four levels of magnification (40$\times$, 100$\times$, 200$\times$ and 400$\times$) are available, which is the case for BreaKHis dataset. However, this algorithm can be applied to any number of malignancy types and magnification levels.

The only parameter passed to the algorithm is the expected number of output cases $k$ (which we assume to be a multiple of the number of types of malignancy). In training set initialization, we want this set size, which we will denote as $k_{\mbox{\scriptsize{\it train}}}$, to be relatively large in order to avoid over-fitting our model later on, but also not too large due to limited computational resources and running time. Therefore, this parameter needs to be fine-tuned for different problem settings as we will show in more detail in Section \ref{sec:experiments}.

Algorithm \ref{algo:casebuild} can be applied to both training and testing sets, depending on the inputs of image sets. Note in the training phase, a case consists of one single image from each magnification level, but not necessary from the same particular patient. The images can be randomly selected from different patients as long as they share the same malignancy. This is why the patient information does not come in Algorithm \ref{algo:casebuild}. However, in the testing phase, we may want the cases to be patient specific, which can be achieved by setting patient specific images as the input to Algorithm \ref{algo:casebuild}. After the whole process, the initialized case-based image sets are ready for training or evaluation.

\begin{algorithm}
	\SetKwData{Left}{left}
	\SetKwData{This}{this}
	\SetKwData{Up}{up}
	\SetKwFunction{Union}{Union}
	\SetKwFunction{FindCompress}{FindCompress}
	\SetKwInOut{Input}{Input}
	\SetKwInOut{Output}{Output}
	\Input{image sets I$_{\mbox{\it Malignancy} \times \mbox{\it Magnification}}$, where {\it Malignancy} is the set of malignancy types, e.g. \{benign, malignant\}, and {\it Magnification} is the set of magnifications, e.g. \{40, 100, 200, 400\}}
	\Output{$X$ $\leftarrow$ data, $y$ $\leftarrow$ label}
	\Parameter{\textit{k} = expected number of output cases}
	\BlankLine
	\emph{Initialize i = 0}\;
	\ForEach{$mal \in \mbox{\it Malignancy}$}{
		\emph{Initialize counter$_{mal}$ = 0} \;
	    \emph{Initialize new current combination $X_i$}\;
		\Repeat{counter$_{mal}$ $\geq$ $k/|\mbox{\it Malignancy}|$}{
			\ForEach{$mag \in \mbox{\it Magnification}$}{
				randomly pick an image from image set I$_\text{\it (mal, mag)}$ and add to $X_i$;
			}
			\If{current combination $X_i$ not in $X$}{
				add $X_i$ to $X$;\\
				$y_i$ = $mal$; \\
				$i$ += 1;\\
				\textit{counter$_{mal}$} += 1;
			}
		}
	}
	\caption{Case-based image set initialization}
	\label{algo:casebuild}
\end{algorithm}

\subsection{ResNet-based classifier}
\label{sec:classifier}
\begin{figure}[ht]
	\centering
	\includegraphics[scale=0.45]{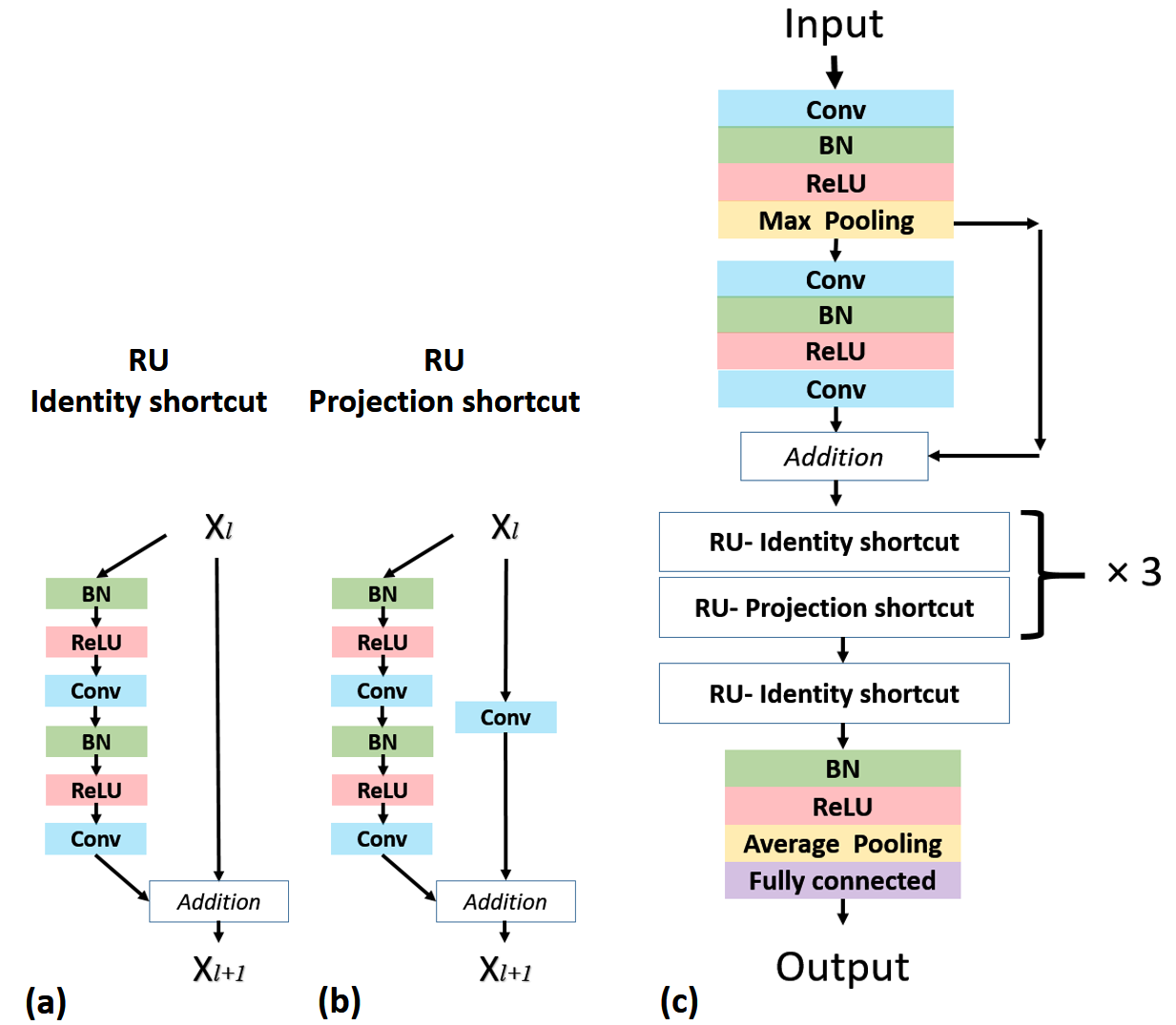}
	\caption{ (a) Residual unit with identity shortcut; (b) Residual unit with projection shortcut; (c) Overall architecture of the ResNet used in the paper. }
	\label{fig:architecture}
\end{figure}

We choose to use deep residual neural networks (ResNets) to classify the histopathological cases. ResNets are a special kind of convolutional neural networks that have residual units in parallel to regular convolutional layers. The design of residual units are quite flexible such that they can also be further engineered in order to get better performance \cite{he2016deep, he2016identity}. We start with a simple 18-layer ResNet model (ResNet-18), as this model can be easily adapted to even deeper models (e.g. 152 layers) if required. The overall architecture of ResNet-18 is shown in Figure \ref{fig:architecture}.

The model contains two types of residual units: the residual unit with an identity shortcut and the residual unit with a projection shortcut. The only difference between these two types is that in the projection shortcut, an additional convolutional layer is required due to the change of dimension from input to output. Each residual unit contains six sequential components: Batch Normalization, Rectified Linear Unit (ReLU), Convolution, Batch Normalization, ReLU and Convolution. An average pooling layer is used before the final fully connected layer.

\subsection{Metrics}
\label{sec:metrics}
Spanhol \etal \cite{spanhol2016breast} have introduced two ways to report method performances for medical image classification: image recognition rate and patient recognition rate. Here, to accommodate for our case-based approach, however, we use a case level metric instead of an image level metric. The case recognition rate is defined as follows:
\begin{equation*}
\text{Case Recognition Rate} = \frac{N_{rec}}{N_{all}}
\tag{1}\label{eq:1}
\end{equation*}
where \textit{N}$_{all}$ is the total number of all cases constructed for the testing set, and \textit{N}$_{rec}$ is the number of correctly classified cases.

Unlike the case recognition rate, the patient recognition rate takes patient information into account. For each patient $p$ in the testing set, let \textit{N}$_{p_{all}}$ be the total number of cases that belong to patient $p$, and \textit{N}$_{p_{rec}}$ be the number of correctly classified cases for patient $p$, then the patient recognition rate can be defined as \cite{spanhol2016breast}:
\begin{equation*}
\text{Patient Recognition Rate} = \frac{\sum_{p} (N_{p_{rec}} / N_{p_{all}})} {\text{Total Number of Patients}}.
\tag{2}\label{eq:2}
\end{equation*}

In addition to the above two recognition rates, we also give a new metric defined at the diagnosis level. First, we give a final diagnosis to each patient in the testing set based on a simple voting strategy where we assume that the diagnosis is benign if the ratio of benign to malignant cases for the patient $p$ is above a threshold, $\mbox{\it malignancy\_threshold}$:
\begin{equation*}
\text{Patient Diagnosis $_{p}$} =\left\{
  \begin{array}{@{}ll@{}}
  benign, & \text{if}\ \frac{N_{p_{benign}}}{N_{p_{all}}} > \mbox{\it malignancy\_threshold} \\
  malignant, & \text{otherwise}
  \end{array}\right.
\tag{3}\label{eq:3}
\end{equation*}
where \textit{N}$_{p_{benign}}$ is the number of cases that are diagnosed as benign for patient $p$. For example, if $\mbox{\it malignancy\_threshold}$ is set to 0.5, the patient $p$ is assigned a diagnosis of benign if more than half of the cases for patient $p$ are classified as benign. Based on the diagnoses assigned to the patients, diagnosis accuracy for the classification is defined as the follows:
\begin{equation*}
\text{Diagnosis Accuracy} = \frac{\text{Number of Correctly Diagnosed Patients}}{\text{Total Number of Patients}}.
\tag{4}\label{eq:4}
\end{equation*}

We believe the diagnosis accuracy metric should be emphasized more for future research on histopathological diagnosis problems, as it is of utmost clinical importance that a computer-aided diagnosis system be able to give a final diagnosis for a patient, and based on the accuracy at which the diagnosis is correct or not, we can judge its performance.

\section{Experiments and Results}
\label{sec:experiments}
This section evaluates our case-based approach for histopathological diagnosis that is proposed in Section \ref{sec:approach}.
\subsection{Dataset}
To test the proposed case-based approach for histopathological diagnosis, we use the BreaKHis database \cite{spanhol2016dataset}, a recently released dataset of breast tumor histopathological images. BreaKHis contains both benign and malignant breast tumor images, which were collected from 82 patients at multiple magnification levels (40$\times$, 100$\times$, 200$\times$ and 400$\times$). Each patient may have a different number of images for each magnification. In total, there are 2480 benign and 5429 malignant images, with each image acquired in three channels (RGB). 

Besides the histopathological images, BreaKHis also provides a five-fold protocol for testing. We use the same testing protocol as previous work \cite{spanhol2016breast, bayramogludeep}, where the whole dataset is split into training (70\%) and testing (30\%) set for five trials, such that none of the images associated with the patients in the training set are used in the testing set. In the end, 54 out of the total of 82 patients are grouped into the training set, and the rest of the 28 patients are used as evaluation samples for all the five folds.

The BreaKHis images are originally of size 700$\times$460$\times$3. To speed up the processing times and lessen the memory requirements, the images are resized to 100$\times$100$\times$3 for both the training and testing sets.

\subsection{Implementation}
With all images from BreaKHis, we implement Algorithm \ref{algo:casebuild} to build histopathological cases for both the training and testing sets. To find the best parameter $k_{\mbox{\scriptsize{\it train}}}$ for Algorithm \ref{algo:casebuild} when initializing the training sets, we utilize fold 1 for a series of experiments by setting the number of output cases over a range of values from 100 to 40,000 as shown in Figure \ref{fig:choosecasenumber}. After comparing the case-level accuracies for the different sizes of training sets, we choose the smallest size of the training set that gives the best accuracy as our final $k_{\mbox{\scriptsize{\it train}}}$. Note that for some of the smaller sizes of the training sets, we repeat some of the experiments independently three or five times since the performance of trained model can vary a lot for these sizes. The final chosen parameter $k_{\mbox{\scriptsize{\it train}}}$ for training set initialization achieves a balance between computational resource requirement and performance. On the other hand, for testing set initialization, we simply use $k_{\mbox{\scriptsize{\it test}}}$ = 30,000 for the size of the testing sets as evaluation on thirty thousands cases gives a quite stable estimation of model performance according to our trial experiments.

For all experiments, we implement our classifier ResNet using Keras, a deep learning library written in python with either TensorFlow or Theano as a backend \cite{chollet2015keras}. We use Theano as the backend in this paper. To optimize the weights, we use stochastic gradient descent, with a batch size of 100 to compute the gradients using back propagation. The initial learning rate is set to 0.001, decay by 1e-6 over each update, and Nestrov momentum is set to 0.9. We train our neural network for 100 epoches. All experiments are done on 4 Intel Xeon(R) E3-1271 v3 processors with a NVIDIA Quadro K2000/PCIe/SSE2 GPU with CUDA 7.5 installed in a Ubuntu 16.04 LTS.

\subsection{Results}
\begin{figure}[ht]
	\centering
	\includegraphics[scale=0.4]{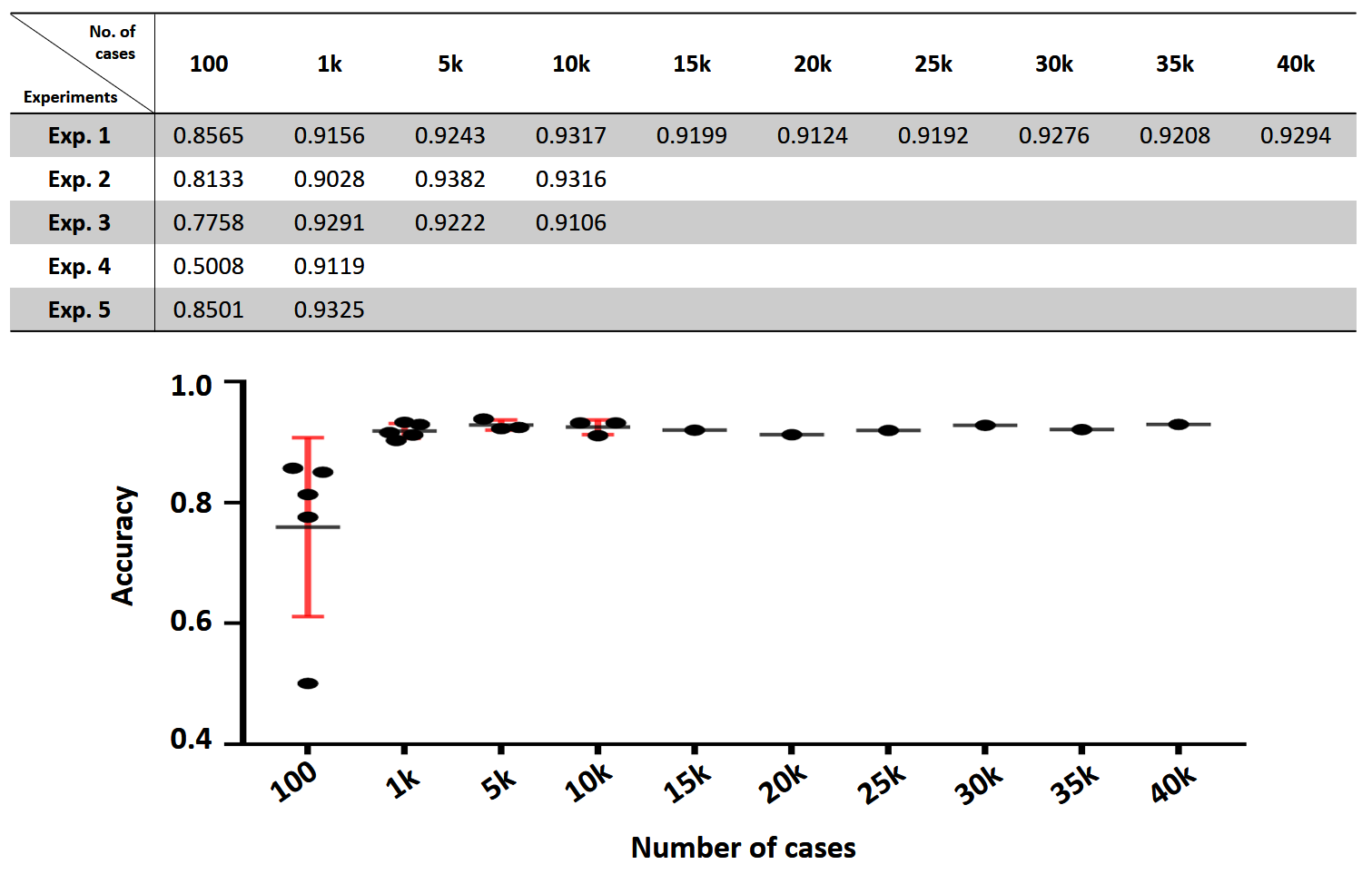}
	\caption{Performance in terms of case-level accuracy versus number of histopathological cases $k_{\mbox{\scriptsize{\it train}}}$ used for training. Top table shows the testing accuracies in each experiment; The bottom plot is the visualization of the table. }
	\label{fig:choosecasenumber}
\end{figure}

First, regarding the choice of $k_{\mbox{\scriptsize{\it train}}}$, as is shown in Figure \ref{fig:choosecasenumber}, with the increase of the total number of cases used for training, the testing accuracy also increases and finally reaches the plateau. When the model is trained on only 100 cases, there is a large variation in model performance based on five independently conducted experiments. In the worst case, for $100$ cases, the performance is not much better than a random guess. However, the model performance is significantly improved when trained on a large number of cases. In addition, the variation becomes smaller as well. From the bottom plot in Figure \ref{fig:choosecasenumber}, we can see that the curve starts to converge when the number of cases is increased to $5,000$, and reaches a maximum at around $10,000$. To understand the effect of the choice of $k_{\mbox{\scriptsize{\it train}}}$ on the running time, when setting $k_{\mbox{\scriptsize{\it train}}}$ equal to $40,000$, the total training time required for a single fold is around 900 seconds per epoch, while it is 2 seconds per epoch for $k_{\mbox{\scriptsize{\it train}}}$ set to $100$. By setting our parameter $k_{\mbox{\scriptsize{\it train}}}$ equal to $10,000$, we significantly reduce the running time by around four times, from 900 seconds to 226 seconds, when compared to $k_{\mbox{\scriptsize{\it train}}}$ equal to $40,000$, without sacrificing accuracy. Therefore, we choose to set our parameter $k_{\mbox{\scriptsize{\it train}}}$ equal to $10,000$ in training set initialization algorithm for all the following experiments.

With the parameters $k_{\mbox{\scriptsize{\it train}}}$ and $k_{\mbox{\scriptsize{\it test}}}$ set, we can then thoroughly evaluate our case-based approach based on five-fold testing protocol, using the metrics that we described in Section \ref{sec:metrics}. For each fold, we first build the case-based training and testing sets using Algorithm \ref{algo:casebuild}, by setting $k_{\mbox{\scriptsize{\it train}}}$ = 10,000 for the training set and $k_{\mbox{\scriptsize{\it test}}}$ = 30,000 for the testing set. Note that both the training and testing sets are balanced in terms of the different malignancy types. After the models are trained, we then evaluate the model performance using the following three metrics: case recognition rate, patient recognition rate, and diagnosis accuracy. For the diagnosis of benign or malignant, we set $\mbox{\it malignancy\_threshold}$ to 0.5. Table \ref{table:accuracies} shows the final results.

\begin{table}[ht]
	\caption{ Performance of case-based approach for histopathological malignancy diagnosis based on case-level, patient-level and diagnosis-level accuracy. }
	\centering
	\begin{tabular}{l C{1.2cm} C{1.0cm} C{1.0cm} C{1.0cm} C{1.0cm} | C{1.8cm}}
		\hline\noalign{\smallskip}
		Accuracy Type & Fold 1 & Fold 2 & Fold 3 & Fold 4 & Fold 5 & Average\\
		\noalign{\smallskip}
		\hline
		\noalign{\smallskip}
		Case Recogn. Rate & 0.9246 & 0.8596 & 0.9355 & 0.9220 & 0.9323 & 0.9148  \\
		Patient Recogn. Rate & 0.8731 & 0.8424 & 0.8753 & 0.8090 & 0.9182 & 0.8636  \\
		Diagnosis Accuracy & $25/28$ & $23/28$ & $26/28$ & $23/28$ & $27/28$ & 0.8857  \\
		\hline
	\end{tabular}
	\label{table:accuracies}
\end{table}

Our case-based approach gives average accuracies of 91.48\% (case-level), 86.36\% (patient-level) and 88.57\% (diagnosis-level) on the testing sets. As we are the first to use case-level and diagnosis-level accuracies, we can't compare the results for these metrics to previous results. However, based on patient-level accuracy, our case-based approach (86.36\%) outperforms the multi-task CNN method (82.13\%, average of four magnifications) \cite{bayramogludeep} and the magnification independent single-task CNN method (83.25\%, average of four magnifications) \cite{bayramogludeep}, and achieves a comparable performance to the best results obtained from the combination of four patch image extraction strategies and three fusion rules using a patch-based method for specific magnifications (40$\times$: 90.0\%; 100$\times$: 88.4\%; 200$\times$: 84.6\%; 400$\times$: 86.1\%) \cite{spanhol2016breast}.

\begin{figure}[ht]
	\centering
	\includegraphics[scale=0.33]{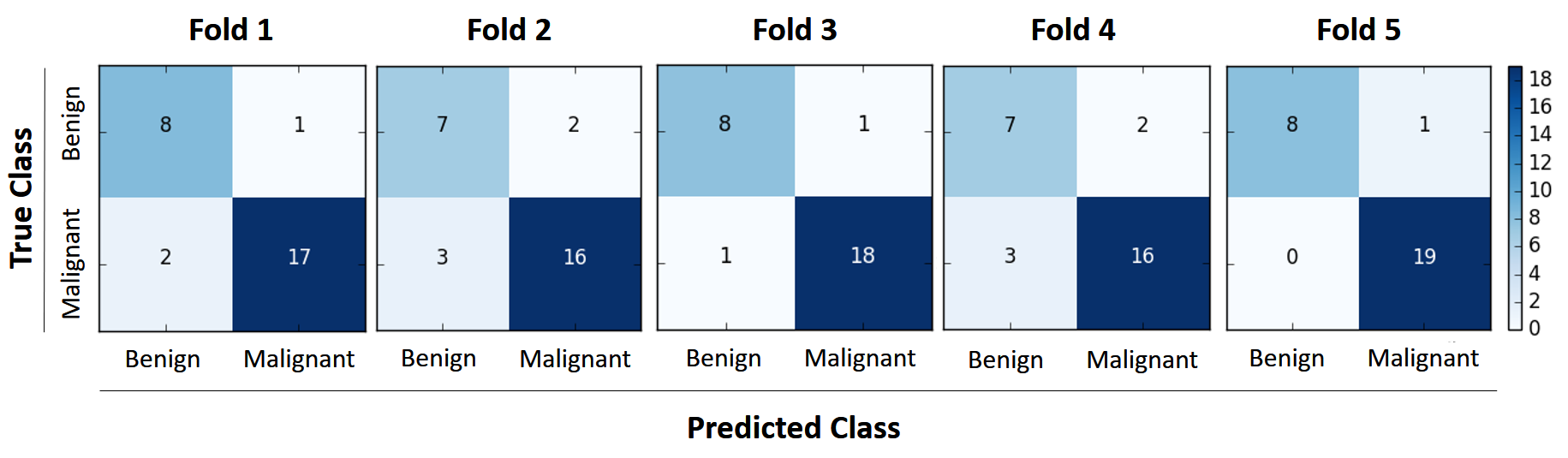}
	\caption{ The confusion matrices of case-based approach for histopathological malignancy diagnosis in five folds. }
	\label{fig:confusionmatrix}
\end{figure}

We further investigate the misclassified patients in terms of malignancy diagnosis for all five folds, and summarize the results as confusion matrices in Figure \ref{fig:confusionmatrix}. In total, 16 out of 140 patient samples over the five folds are misclassified, with a false positive rate of 5.0\% and a false negative rate of 6.43\%.

\section{Conclusion}
In this paper, we propose a case-based approach for histopathological malignancy diagnosis using deep residual neural networks. We first introduce an algorithm for case-based image set initialization for both training and testing based on histopathological images at multiple magnification levels, and then present a ResNet-based classifier and three metrics to report method performances for medical image classification. Finally, we evaluate our proposed approach using the breast tumor histopathological image dataset BreaKHis. Our results show that the case-based approach achieves better performance than the state-of-the-art methods. Moreover, we believe our case-based approach is a more reasonable way for histopathological malignancy classification since it makes diagnosis decision based on features learned at multiple magnifications. Another principle advantage of our work over the previous work \cite{spanhol2016breast, bayramogludeep} is that our method gives a single diagnosis for the patient, whereas in the previous work four potentially differing diagnoses are given for the same patient, one for each of four magnification levels. To be clinically applicable, these latter approaches would then require a final voting step or similar diagnosis selection step which are not discussed in their papers \cite{spanhol2016breast, bayramogludeep}. For future work, more complex deep CNN architectures will be investigated.

\bibliography{egbib}

\begin{thebibliography}{21}
\providecommand{\natexlab}[1]{#1}
\providecommand{\url}[1]{\texttt{#1}}
\expandafter\ifx\csname urlstyle\endcsname\relax
  \providecommand{\doi}[1]{doi: #1}\else
  \providecommand{\doi}{doi: \begingroup \urlstyle{rm}\Url}\fi

\bibitem[Bayramoglu et~al.(2016)Bayramoglu, Kannala, and
  Heikkil{\"a}]{bayramogludeep}
Neslihan Bayramoglu, Juho Kannala, and Janne Heikkil{\"a}.
\newblock Deep learning for magnification independent breast cancer
  histopathology image classification.
\newblock In \emph{23th International Conference on Pattern Recognition, ICPR},
  2016.

\bibitem[Chollet(2015)]{chollet2015keras}
Fran\c{c}ois Chollet.
\newblock Keras, 2015.
\newblock \url{https://github.com/fchollet/keras}.

\bibitem[Cire{\c{s}}an et~al.(2013)Cire{\c{s}}an, Giusti, Gambardella, and
  Schmidhuber]{cirecsan2013mitosis}
Dan~C Cire{\c{s}}an, Alessandro Giusti, Luca~M Gambardella, and J{\"u}rgen
  Schmidhuber.
\newblock Mitosis detection in breast cancer histology images with deep neural
  networks.
\newblock In \emph{International Conference on Medical Image Computing and
  Computer-assisted Intervention}, pages 411--418. Springer, 2013.

\bibitem[Cruz-Roa et~al.(2014)Cruz-Roa, Basavanhally, Gonz{\'a}lez, Gilmore,
  Feldman, Ganesan, Shih, Tomaszewski, and Madabhushi]{cruz2014automatic}
Angel Cruz-Roa, Ajay Basavanhally, Fabio Gonz{\'a}lez, Hannah Gilmore, Michael
  Feldman, Shridar Ganesan, Natalie Shih, John Tomaszewski, and Anant
  Madabhushi.
\newblock Automatic detection of invasive ductal carcinoma in whole slide
  images with convolutional neural networks.
\newblock In \emph{SPIE medical imaging}, pages 904103--904103. International
  Society for Optics and Photonics, 2014.

\bibitem[Demir and Yener(2005)]{demir2005automated}
Cigdem Demir and B{\"u}lent Yener.
\newblock Automated cancer diagnosis based on histopathological images: a
  systematic survey.
\newblock \emph{Rensselaer Polytechnic Institute, Tech. Rep}, 2005.

\bibitem[Gurcan et~al.(2009)Gurcan, Boucheron, Can, Madabhushi, Rajpoot, and
  Yener]{gurcan2009histopathological}
Metin~N Gurcan, Laura~E Boucheron, Ali Can, Anant Madabhushi, Nasir~M Rajpoot,
  and Bulent Yener.
\newblock Histopathological image analysis: A review.
\newblock \emph{IEEE reviews in biomedical engineering}, 2:\penalty0 147--171,
  2009.

\bibitem[He et~al.(2016{\natexlab{a}})He, Zhang, Ren, and Sun]{he2016deep}
Kaiming He, Xiangyu Zhang, Shaoqing Ren, and Jian Sun.
\newblock Deep residual learning for image recognition.
\newblock In \emph{Proceedings of the IEEE Conference on Computer Vision and
  Pattern Recognition}, pages 770--778, 2016{\natexlab{a}}.

\bibitem[He et~al.(2016{\natexlab{b}})He, Zhang, Ren, and Sun]{he2016identity}
Kaiming He, Xiangyu Zhang, Shaoqing Ren, and Jian Sun.
\newblock Identity mappings in deep residual networks.
\newblock In \emph{European Conference on Computer Vision}, pages 630--645.
  Springer, 2016{\natexlab{b}}.

\bibitem[He et~al.(2012)He, Long, Antani, and Thoma]{he2012histology}
Lei He, L~Rodney Long, Sameer Antani, and George~R Thoma.
\newblock Histology image analysis for carcinoma detection and grading.
\newblock \emph{Computer methods and programs in biomedicine}, 107\penalty0
  (3):\penalty0 538--556, 2012.

\bibitem[Krizhevsky et~al.(2012)Krizhevsky, Sutskever, and
  Hinton]{krizhevsky2012imagenet}
Alex Krizhevsky, Ilya Sutskever, and Geoffrey~E Hinton.
\newblock Imagenet classification with deep convolutional neural networks.
\newblock In \emph{Advances in neural information processing systems}, pages
  1097--1105, 2012.

\bibitem[LeCun et~al.(1989)LeCun, Boser, Denker, Henderson, Howard, Hubbard,
  and Jackel]{lecun1989backpropagation}
Yann LeCun, Bernhard Boser, John~S Denker, Donnie Henderson, Richard~E Howard,
  Wayne Hubbard, and Lawrence~D Jackel.
\newblock Backpropagation applied to handwritten zip code recognition.
\newblock \emph{Neural computation}, 1\penalty0 (4):\penalty0 541--551, 1989.

\bibitem[LeCun et~al.(2015)LeCun, Bengio, and Hinton]{lecun2015deep}
Yann LeCun, Yoshua Bengio, and Geoffrey Hinton.
\newblock Deep learning.
\newblock \emph{Nature}, 521\penalty0 (7553):\penalty0 436--444, 2015.

\bibitem[Litjens et~al.(2016)Litjens, S{\'a}nchez, Timofeeva, Hermsen,
  Nagtegaal, Kovacs, Hulsbergen-Van De~Kaa, Bult, Van~Ginneken, and Van
  Der~Laak]{litjens2016deep}
Geert Litjens, Clara~I S{\'a}nchez, Nadya Timofeeva, Meyke Hermsen, Iris
  Nagtegaal, Iringo Kovacs, Christina Hulsbergen-Van De~Kaa, Peter Bult, Bram
  Van~Ginneken, and Jeroen Van Der~Laak.
\newblock Deep learning as a tool for increased accuracy and efficiency of
  histopathological diagnosis.
\newblock \emph{Scientific reports}, 6, 2016.

\bibitem[McCann et~al.(2015)McCann, Ozolek, Castro, Parvin, and
  Kovacevic]{mccann2015automated}
Michael~T McCann, John~A Ozolek, Carlos~A Castro, Bahram Parvin, and Jelena
  Kovacevic.
\newblock Automated histology analysis: Opportunities for signal processing.
\newblock \emph{IEEE Signal Processing Magazine}, 32\penalty0 (1):\penalty0
  78--87, 2015.

\bibitem[Roa-Pe{\~n}a et~al.(2010)Roa-Pe{\~n}a, G{\'o}mez, and
  Romero]{roa2010experimental}
Lucia Roa-Pe{\~n}a, Francisco G{\'o}mez, and Eduardo Romero.
\newblock An experimental study of pathologist's navigation patterns in virtual
  microscopy.
\newblock \emph{Diagnostic pathology}, 5\penalty0 (1):\penalty0 71, 2010.

\bibitem[Romo et~al.(2014)Romo, Garc{\'\i}a-Arteaga, Arbel{\'a}ez, and
  Romero]{romo2014discriminant}
David Romo, Juan~D Garc{\'\i}a-Arteaga, Pablo Arbel{\'a}ez, and Eduardo Romero.
\newblock A discriminant multi-scale histopathology descriptor using dictionary
  learning.
\newblock In \emph{SPIE Medical Imaging}, pages 90410Q--90410Q. International
  Society for Optics and Photonics, 2014.

\bibitem[Rubin et~al.(2008)Rubin, Strayer, Rubin, et~al.]{rubin2008rubin}
Raphael Rubin, David~S Strayer, Emanuel Rubin, et~al.
\newblock \emph{Rubin's pathology: clinicopathologic foundations of medicine}.
\newblock Lippincott Williams \& Wilkins, 2008.

\bibitem[Sirinukunwattana et~al.(2016)Sirinukunwattana, Raza, Tsang, Snead,
  Cree, and Rajpoot]{sirinukunwattana2016locality}
Korsuk Sirinukunwattana, Shan E~Ahmed Raza, Yee-Wah Tsang, David~RJ Snead,
  Ian~A Cree, and Nasir~M Rajpoot.
\newblock Locality sensitive deep learning for detection and classification of
  nuclei in routine colon cancer histology images.
\newblock \emph{IEEE transactions on medical imaging}, 35\penalty0
  (5):\penalty0 1196--1206, 2016.

\bibitem[Spanhol et~al.(2016{\natexlab{a}})Spanhol, Oliveira, Petitjean, and
  Heutte]{spanhol2016dataset}
Fabio~A Spanhol, Luiz~S Oliveira, Caroline Petitjean, and Laurent Heutte.
\newblock A dataset for breast cancer histopathological image classification.
\newblock \emph{IEEE Transactions on Biomedical Engineering}, 63\penalty0
  (7):\penalty0 1455--1462, 2016{\natexlab{a}}.

\bibitem[Spanhol et~al.(2016{\natexlab{b}})Spanhol, Oliveira, Petitjean, and
  Heutte]{spanhol2016breast}
Fabio~Alexandre Spanhol, Luiz~S Oliveira, Caroline Petitjean, and Laurent
  Heutte.
\newblock Breast cancer histopathological image classification using
  convolutional neural networks.
\newblock In \emph{2016 International Joint Conference on Neural Networks
  (IJCNN)}, pages 2560--2567. IEEE, 2016{\natexlab{b}}.

\bibitem[Veta et~al.(2014)Veta, Pluim, Van~Diest, and
  Viergever]{veta2014breast}
Mitko Veta, Josien~PW Pluim, Paul~J Van~Diest, and Max~A Viergever.
\newblock Breast cancer histopathology image analysis: A review.
\newblock \emph{IEEE Transactions on Biomedical Engineering}, 61\penalty0
  (5):\penalty0 1400--1411, 2014.

\end{thebibliography}
\end{document}